\documentclass[11pt,a4paper]{article}
\usepackage[hyperref]{acl2020}
\usepackage{times}
\usepackage{latexsym}

\usepackage{microtype}

\aclfinalcopy 
\def\aclpaperid{2281} 

\usepackage{siunitx}
\usepackage{array}
\usepackage{multirow}
\usepackage{makecell}
\usepackage{amsmath}
\usepackage{verbatim}
\usepackage{amsfonts}
\usepackage{graphicx}
\usepackage{color}
\usepackage{float}
\usepackage{bm}
\usepackage{arydshln}
\usepackage[noend]{algpseudocode}
\usepackage{tikz}
\usepackage{booktabs}
\usepackage[linesnumbered, ruled, vlined]{algorithm2e}
\newcommand*{\Scale}[2][4]{\scalebox{#1}{$#2$}}  

\title{Few-shot Slot Tagging with Collapsed Dependency Transfer and Label-enhanced Task-adaptive Projection Network}

\author{
	Yutai Hou$^{1}$,
	Wanxiang Che$^{1}$\thanks{$\ $ Corresponding author.}~,
	Yongkui Lai$^{1}$,
	Zhihan Zhou$^{3}$, \\
	\textbf{Yijia Liu}$^{2}$,
	\textbf{Han Liu}$^{3}$,
	\textbf{Ting Liu}$^{1}$
	\\
	$^{1}$Research Center for Social Computing and Information Retrieval, \\ Harbin Institute of Technology \\
	$^{2}$Alibaba Group \quad $^{3}$Department of Computer Science, Northwestern University  \\
	{\tt \{ythou, car, yklai, tliu\}@ir.hit.edu.cn,} {\tt oneplus.lau@gmail.com} \\
	{\tt zhihanzhou2020@u.northwestern.edu, hanliu@northwestern.edu} \\
}

\date{}

\begin{document}
\maketitle
\begin{abstract}
In this paper, we explore the
slot tagging with only a few labeled support sentences (a.k.a. few-shot).
Few-shot slot tagging faces a 
unique challenge compared to the other
few-shot classification problems as it calls for modeling the dependencies between labels.
But it is hard to apply previously learned label dependencies to an unseen domain, 
due to the discrepancy of label sets.
To tackle this, we introduce a \textit{collapsed dependency transfer} mechanism into the conditional random field (CRF) to transfer abstract label dependency patterns as transition scores.
In the few-shot setting, 
the emission score of CRF can be calculated as a word's similarity to the representation of each label. 
To calculate such similarity, 
we propose a \textit{Label-enhanced Task-Adaptive Projection Network (L-TapNet)} based on 
the state-of-the-art few-shot classification model -- TapNet, 
by leveraging label name semantics in representing labels. 
Experimental results show that
our model significantly outperforms the strongest few-shot learning baseline by 
14.64 F1 scores in the one-shot setting.\footnote{Code is available at: \url{https://github.com/AtmaHou/FewShotTagging}}
\end{abstract} 

\section{Introduction}
Slot tagging \cite{tur2011spoken}, 
a key module in the task-oriented 
dialogue system \cite{young2013pomdp}, 
is usually formulated as a sequence 
labeling problem \cite{sarikaya2016overview}. 
Slot tagging faces the rapid changing 
of domains, 
and the labeled data
is usually scarce for new domains with only a few samples. 
Few-shot learning technique \cite{miller2000learning,fei2006one,lake2015human,matching}
is appealing in this scenario 
since it learns the model 
that borrows the prior experience 
from old domains and adapts to new domains quickly with only very few examples
(usually one or two examples for each class).
	
\begin{figure}[t]
	\centering
	\begin{tikzpicture}
	\draw (0,0 ) node[inner sep=0] {\includegraphics[width=1\columnwidth, trim={18.5cm 13cm 12.5cm 4.8cm}, clip]{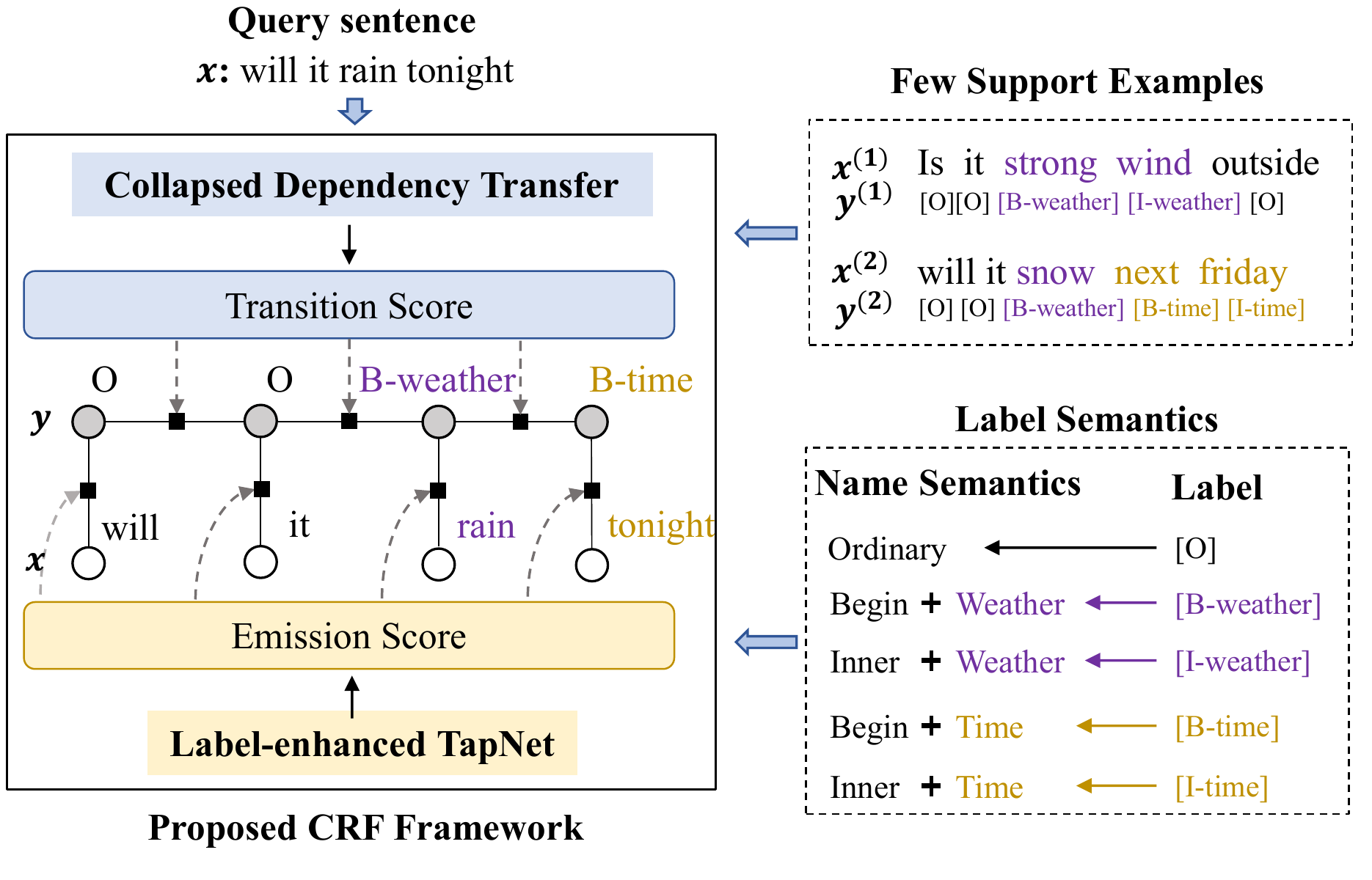}};
	\end{tikzpicture}  
	\caption{\footnotesize
		Our few-shot CRF framework for slot tagging. 
	}\label{fig:intro1}
	\vspace*{-4mm}
\end{figure}

Previous few-shot learning studies mainly focused on classification problems,
which have been widely explored with similarity-based methods \citep{matching,prototypical,sung2018learning,yan2018few,yu2018diverse}. 
The basic idea of these methods is 
classifying an (query) item
in a new domain  
according to its similarity with the representation of each 
class.
The similarity function is usually learned in prior rich-resource domains and 
per class representation is obtained from few labeled samples (support set).
It is straight-forward to 
decompose the few-shot sequence labeling into a series of independent few-shot classifications
and apply the similarity-based methods.
However, 
sequence labeling benefits from taking the dependencies between labels into account \cite{huang2015bidirectional,ma2016end}.
To consider both the item similarity and label dependency,
we propose to 
leverage the conditional random fields
 \citep[CRFs]{CRF} 
 in few-shot sequence labeling (see Figure \ref{fig:intro1}).
In this paper,
we translate the emission score of CRF into the output of the similarity-based method 
and calculate the transition score with a specially designed transfer mechanism.

The few-shot scenario poses unique challenges in learning the emission and transition scores of CRF.
It is infeasible to learn the transition on the few labeled data,
and prior label dependency in source domain cannot be directly transferred
due to discrepancy in label set.
To tackle the label discrepancy problem, 
we introduce the \textit{collapsed dependency transfer} mechanism. 
It transfers label dependency information from source domains to target domains 
by abstracting domain-specific labels into abstract domain-independent labels and modeling the label dependencies between these abstract labels.

It is also challenging to compute the emission scores
(word-label similarity in our case). 
Popular few-shot models, such as Prototypical Network \cite{prototypical}, 
average the embeddings of each label's support examples as label representations, 
which often distribute closely in the embedding space and thus cause misclassification.
To remedy this, \citet{yoon2019tapnet} propose TapNet that 
learns to project embedding to a space where words of different labels are well-separated. 
We introduce this idea to slot tagging and further propose to improve label representation by leveraging the semantics of label names. 
We argue that label names are often semantically related to slot words 
and can help word-label similarity modeling.
For example in Figure \ref{fig:intro1}, 
word \textit{rain} and label name \textit{weather} are highly related.
To use label name semantic and achieve good-separating in label representation, 
we propose \textit{Label-enhanced TapNet} (L-TapNet) that constructs an embedding projection space using label name semantics, 
where label representations are well-separated and aligned with embeddings of both label name and slot words. 
Then we calculate similarities in the projected embedding space. 
Also, we introduce a \textit{pair-wise embedding} mechanism to representation words with domain-specific context. 

One-shot and five-shot experiments on slot tagging and named entity recognition show that our model
achieves significant improvement over the strong few-shot learning baselines. 
Ablation tests demonstrate improvements coming from both L-TapNet and collapsed dependency transfer. 
Further analysis for label dependencies shows it captures non-trivial information and outperforms transition based on rules. 

\begin{figure*}[t]
	\centering
	\begin{tikzpicture}
	\draw (0,0 ) node[inner sep=0] {\includegraphics[width=2.08\columnwidth, trim={5.5cm 6.3cm 12.35cm 3.6cm}, clip]{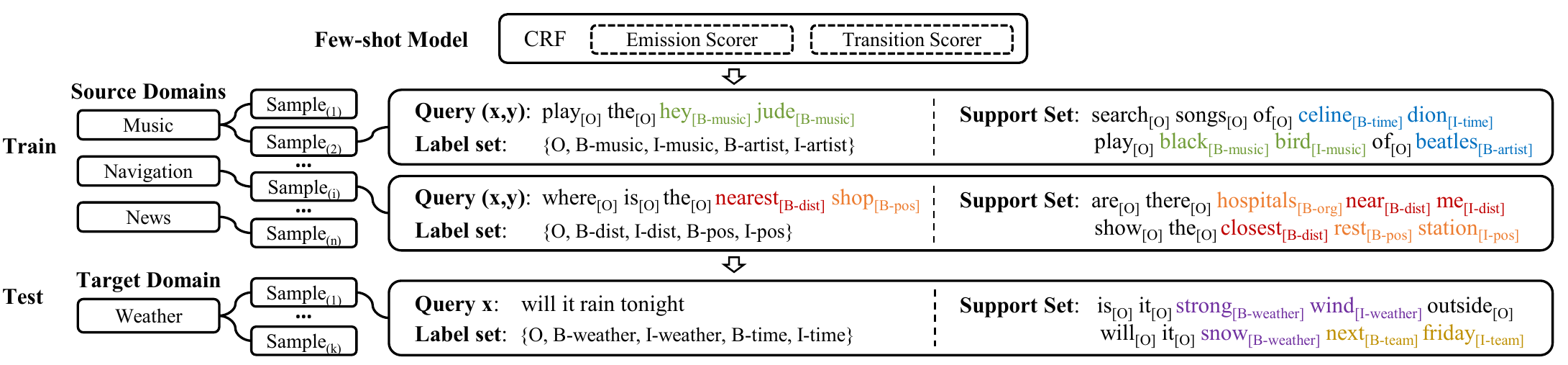}};
	\end{tikzpicture}
	\caption{\footnotesize
		Overviews of training and testing.
		This figure illustrates the procedure of training the model on a set of source domains, and testing it on an unseen domain with only a support set. }\label{fig:overview}
		\vspace*{-4mm}
\end{figure*}

Our contributions are summarized as follows:
(1) We propose a few-shot CRF framework for slot tagging that computes emission score as word-label similarity and estimate transition score by transferring previously learned label dependencies. 
(2) We introduce the collapsed dependency transfer mechanism to transfer label dependencies across domains with different label sets. 
(3) We propose the L-TapNet to leverage semantics of label names to enhance label representations, which help to model the word-label similarity.

\section{Problem Definition}\label{sec:p_def}
We define sentence $\bm{x} = (x_1, x_2, \ldots, x_n)$ as a sequence of words 
and define label sequence of the sentence as $\bm{y} = (y_1, y_2, \ldots, y_n)$.
A domain $\mathcal{D}  = \left\{(\bm{x}^{(i)},\bm{y}^{(i)})\right\}_{i=1}^{N_D}$ is a set of $(\bm{x},\bm{y})$ pairs. 
For each domain, 
there is a corresponding domain-specific label set 
$\mathcal{L_D} =\left\{ \ell_i \right\}_{i=1}^{N}$. 
To simplify the description, 
we assume that 
the number of labels $N$ is same for all domains.

As shown in Figure \ref{fig:overview}, few-shot models are usually first trained on a set of source domains $\left\{\mathcal{D}_1, \mathcal{D}_2, \ldots \right\}$, 
then directly work on another set of unseen target domains $\left\{\mathcal{D}_1', \mathcal{D}_2', \ldots \right\}$ 
without fine-tuning. 
A target domain $\mathcal{D}_j'$ only contains few labeled samples,
which is called support set $\mathcal{S} = \left\{(\bm{x}^{(i)},\bm{y}^{(i)})\right\}_{i=1}^{N_\mathcal{S}}$.
$\mathcal{S}$ usually includes $k$ examples (K-shot) for each of $N$ labels (N-way).

The K-shot sequence labeling task is defined as follows: 
given a K-shot support set $\mathcal{S}$ and an input query sequence $\bm{x} = (x_1, x_2, \ldots, x_n)$, 
find $\bm{x}$'s best label sequence $\bm{y}^*$:
\[
\Scale[0.85]{
\bm{y}^* = (y_1, y_2, \ldots, y_n) = \mathop{\arg\max}_{\bm{y}} \ \ p(\bm{y}  \mid  \bm{x}, \mathcal{S})
}.
\]

\section{Model}\label{sec:model}
In this section, we first show the overview of the proposed CRF framework (\S\ref{sec:overview}). Then we discuss how to compute label transition score with collapsed dependency transfer (\S\ref{sec:trans}) and compute emission score with L-TapNet (\S\ref{sec:emission}). 

\subsection{Framework Overview}\label{sec:overview}
Conditional Random Field (CRF) considers both the transition score and the emission score to find the global optimal label sequence for each input.
Following the same idea, 
we build our few-shot slot tagging framework with two components: 
Transition Scorer and Emission Scorer. 


We apply the linear-CRF to the few-shot setting 
by modeling the label probability of label $\bm{y}$ given query sentence $\bm{x}$ and a K-shot support set $\mathcal{S}$:
\[
\Scale[0.85]{
p(\bm{y} \mid \bm{x}, \mathcal{S}) = \frac{1}{Z}\exp(\text{TRANS}(\bm{y}) + \lambda \cdot \text{EMIT}(\bm{y}, \bm{x}, \bm{S}))
},
\]
where 
$
\Scale[0.85]{
Z = \displaystyle \sum_{\bm{y}' \in \bm{Y}}\exp (\text{TRANS}(\bm{y}') + \lambda \cdot \text{EMIT}(\bm{y}', \bm{x}, \bm{S}))
}
$, 
$\Scale[0.85]{\text{TRANS}(\bm{y})=\sum_{i=1}^n f_T(y_{i-1}, y_i)}$ is the Transition Scorer output and $\Scale[0.85]{\text{EMIT}(\bm{y}, \bm{x}, \bm{S}) = \sum_{i=0}^n f_E(y_i, \bm{x}, \mathcal{S})}$ is the Emission Scorer output. 
$\lambda$ is a scaling parameter which balances weights of the two scores.

We take $L_{\text{CRF}} = -\log (p(\bm{y} \mid \bm{x}, \mathcal{S}))$ as loss function  
and minimize it on data from source domains. 
After the model is trained, we employ Viterbi algorithm 
\citep{viterbi} to find the best label sequence for each input. 


\subsection{Transition Scorer}\label{sec:trans}
The transition scorer component captures the dependencies between labels.\footnote{Here, we ignore $Start$ and $End$ labels for simplicity. In practice, $Start$ and $End$ are included as two additional abstract labels.} 
We model the label dependency as the transition probability between two labels:
\vspace*{-2mm}
\[
\Scale[0.85]{
    f_T(y_{i-1}, y_i) = p(y_i \mid y_{i-1} ) 
}.
\]

Conventionally, such probabilities are learned from training data and stored in a transition matrix $\bm{T}^{N \times N}$, where $N$ is the number of labels. 
For example, $\bm{T}_{\text{B-loc},\text{B-team}}$ corresponds to 
$p(\text{B-loc} \mid \text{B-team})$.
But in the few-shot setting, a model faces different label sets 
in the source domains (train) and the target domains (test). 
This mismatch on labels blocks the trained transition scorer
directly working on a target domain.

\paragraph{Collapsed Dependency Transfer Mechanism}
\begin{figure}[t]
	\centering 
	\begin{tikzpicture}
	\draw (0,0 ) node[inner sep=0] {\includegraphics[width=1\columnwidth, trim={11.2cm 7.5cm 9.7cm 7.6cm}, clip]{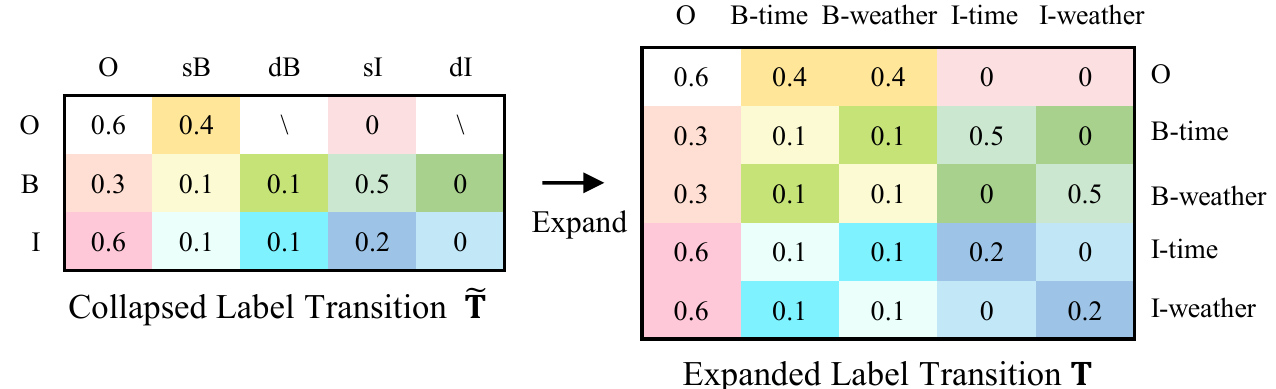}};
	\end{tikzpicture}
	\caption{\footnotesize
	    An example of collapsed label dependency transfer.  
	    We learn a collapsed label transition $\bm{\tilde{T}}$ and obtain specific label transition $\bm{T}$ by filling each position of it with value from $\bm{\tilde{T}}$ in the same color.
	}\label{fig:trans}
\vspace*{-3mm}
\end{figure}
We overcome the above issue by  
directly modeling the transition probabilities between abstract labels.
Intuitively, we collapse specific labels into three abstract labels:  $O$, $B$ and $I$. 
To distinguish whether two labels are under the same or different semantics, 
we model transition from $B$ and $I$ to the same $B$ ($sB$), a different $B$ ($dB$), the same $I$ ($sI$) and a different $I$ ($dI$).
We record such abstract label transition with a Table $\bm{\tilde{T}}^{3\times5}$ (see Figure \ref{fig:trans}).
For example, $\bm{\tilde{T}}_{B,sB}=p(B \text{-} {\ell_m} \mid B\text{-}{\ell_m} )$ is the transition probability of two same $B$ labels. 
And $\bm{\tilde{T}}_{B,dI} = p(I\text{-}{\ell_n} \mid B\text{-}{\ell_m} )$ is the transition probability from a $B$ label to an $I$ label with different types, 
where $\ell_m \neq \ell_n$.
$\bm{\tilde{T}}_{O,sB}$ and $\bm{\tilde{T}}_{O,sI}$ respectively stands for 
the probability of transition from $O$ to any $B$ or $I$ label. 

To calculate the label transition probability for a new domain, 
we construct the transition matrix $\bm{T}$ by filling it with values in $\bm{\tilde{T}}$.
Figure \ref{fig:trans} shows the filling process,
where positions in the same color are filled by the same values.
For example, 
we fill $\bm{T}_{\text{B-loc},\text{B-team}}$ with value in $\bm{\tilde{T}}_{B,dB}$.

\subsection{Emission Scorer}\label{sec:emission}
\begin{figure}[t]
	\centering 
	\begin{tikzpicture}
	\draw (0,0 ) node[inner sep=0] {\includegraphics[width=1\columnwidth, trim={4.6cm 3.7cm 15.5cm 8.7cm}, clip]{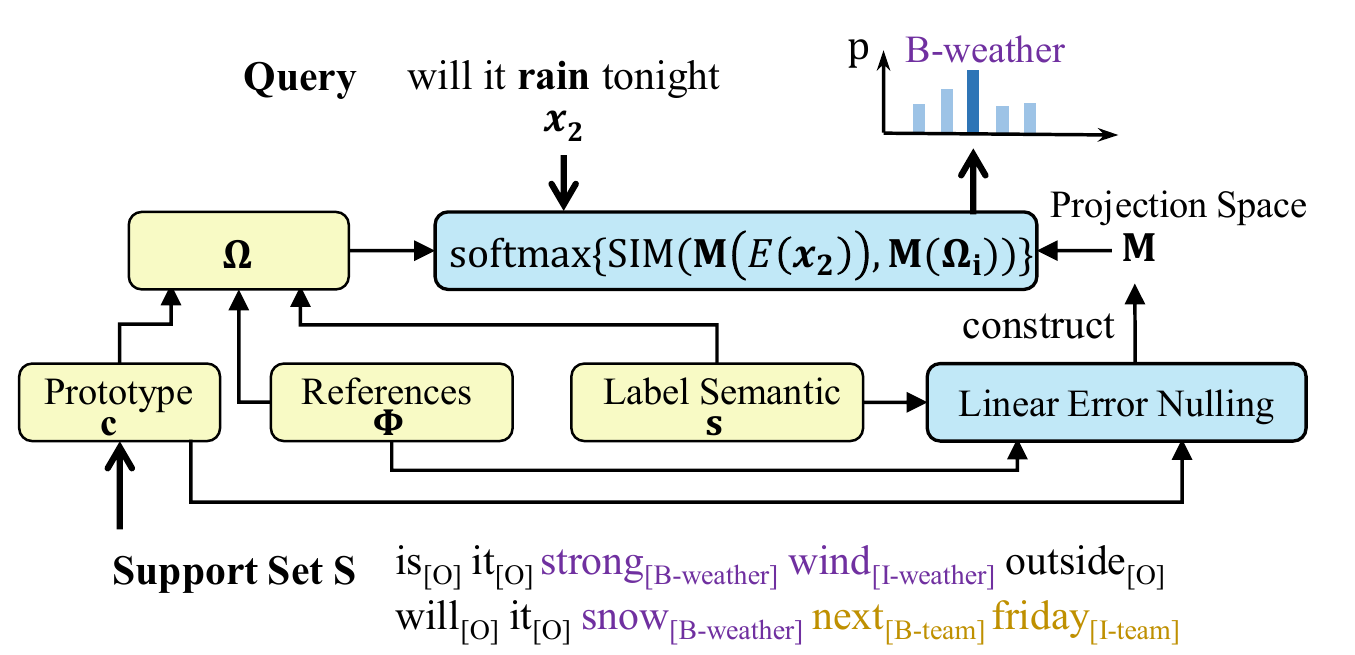}};
	\end{tikzpicture}
	\caption{\footnotesize
	    Emission Scorer with L-TapNet.  
		It first constructs a projection space $\bm{M}$ by \textit{linear error nulling} for given domain, 
		and then predicts a word's emission score with its distance to label representation $\bm{\Omega}$ in the projection space. 
	}\label{fig:ems}
\vspace*{-1mm}
\end{figure}

As shown in Figure \ref{fig:ems}, the emission scorer 
independently assigns each word an emission score with regard to each label: 
\vspace*{-2mm}
\[
\Scale[0.85]{
	f_E(y_i, \bm{x}, \mathcal{S}) = p(y_i \mid \bm{x}, \mathcal{S})
}.
\]

In few-shot setting, 
a word's emission score is calculated according to its similarity to representations of each label. 
To compute such emission, 
we propose the L-TapNet by improving TapNet \cite{yoon2019tapnet} with label semantics and prototypes.   


\subsubsection{Task-Adaptive Projection Network}
TapNet is the state-of-the-art few-shot image classification model. 
Previous few-shot models, 
such as Prototypical Network, 
average the embeddings of each label’s support example as label representations and 
directly compute word-label similarity in word embedding space. 
Different from them, 
TapNet calculates word-label similarity in a projected embedding space, 
where the words of different labels are well-separated. 
That allows TapNet to reduce misclassification. 
To achieve this, 
TapNet leverages a set of per-label reference vectors $\mathbf{\Phi}=[\boldsymbol{\phi}_{1};\cdots;\boldsymbol{\phi}_{N}]$ as label representations.  
and construct a projection space based on these references. 
Then, a word $x$'s emission score for label $\ell_j$ is calculated as its similarity to reference $\boldsymbol{\phi}_j$:
\[
\Scale[0.85]{
\begin{array}{l}
	f_E(y_j, \bm{x}, \mathcal{S})  = \text{Softmax}\{\textsc{Sim}(\mathbf{M}( E(x)), \mathbf{M}(\boldsymbol{\phi}_{j}) \}
\end{array}
},
\]
where $\mathbf{M}$ is a projecting function, $E$ is an embedder and $\textsc{Sim}$ is a similarity function. 
TapNet shares the references $\mathbf{\Phi}$ across different domains 
and constructs $\mathbf{M}$ for each specific domain by randomly associating the references to the specific labels.

\paragraph{Task-Adaptive Projection Space Construction} 
Here, we present a brief introduction for the construction of projection space. 
Let $\mathbf{c}_{j}$ be the average of the embedded features for words 
with label $\ell_j$ in support set $S$. 
Given the $\mathbf{\Phi}=[\boldsymbol{\phi}_{1};\cdots;\boldsymbol{\phi}_{N}]$ and support set $S$, 
TapNet constructs the projector $\mathbf{M}$ such that (1) each $\mathbf{c}_{j}$ and corresponding reference vector $\boldsymbol{\phi}_j$  align closely when projected by $\mathbf{M}$. (2) words of different labels are well-separated when projected by $\mathbf{M}$. 

To achieve these, 
TapNet first computes the alignment bias between $\mathbf{c}_{j}$ and $\boldsymbol{\phi}_j$ in original embedding space,
then it finds a projection $\mathbf{M}$ that eliminates this alignment bias and effectively separates different labels at the same time.
Specifically, TapNet takes the matrix solution of a \texttt{linear error nulling} process as the embedding projector $\mathbf{M}$.
For the detail process, refer to the original paper.

\subsubsection{Label-enhanced TapNet}
As mentioned in the introduction, 
we argue that label names often semantically relate to slot words 
and can help word-label similarity modeling.
To enhance TapNet with such information,
we use label semantics in both label representation and construction of projection space.

\paragraph{Projection Space with Label Semantics}
Let prototype $\mathbf{c}_{j}$ be the average of embeddings of words with label $\ell_j$ in support set. 
And $\mathbf{s}_j$ is semantic representation of label $\ell_j$ and Section~\ref{sec:emb} will introduce how to obtain it in detail.
Intuitively, slot values ($\mathbf{c}_{j}$) and corresponding label name ($\boldsymbol{s}_j$) often have related semantics and they should be close in embedding space. 
So, we find a projector $\mathbf{M}$ that aligns $\mathbf{c}_{j}$ to both $\boldsymbol{\phi}_j$ and $\mathbf{s}_j$.
The difference with TapNet is that it only aligns $\mathbf{c}_{j}$ to references $\boldsymbol{\phi}_j$ but we also require alignments with label representation. 
The label-enhanced reference is calculated as:
\[
\Scale[0.85]{
    \small
    \boldsymbol{\psi}_j = (1 - \alpha) \cdot \boldsymbol{\phi}_j + \alpha \mathbf{s}_j
},
\]
where $\alpha$ is a balance factor. 
Label semantics $\mathbf{s}_j$ makes $\mathbf{M}$ specific for each domain.
And reference  $\boldsymbol{\phi}_j$ provides cross domain generalization.  

Then we construct an $\mathbf{M}$ by linear error nulling of alignment error between label enhanced reference $\boldsymbol{\psi}_j$ and  $\mathbf{c}_{j}$ following the same steps of TapNet.

\paragraph{Emission Score with Label Semantic}
For emission score calculation, 
compared to TapNet that only uses domain-agnostic reference $\boldsymbol{\phi}$ as label representation, 
we also consider the label semantics and use the label-enhanced reference $\boldsymbol{\psi}_j$ in label representation. 

Besides, 
we further incorporate the idea of Prototypical Network and 
represent a label using a \textbf{prototype reference} $\mathbf{c}_{j}$ as  $\boldsymbol{\Omega}_j = (1 - \beta) \cdot \mathbf{c}_{j} + \beta \boldsymbol{\psi}_j$. 
Finally, the emission score of $x$ is calculated as its similarity to label representation $\boldsymbol{\Omega}$:
\[
\Scale[0.85]{
\begin{array}{l}
	f_E(y_j, \bm{x}, \mathcal{S}) = \text{Softmax}\{\textsc{Sim}(\mathbf{M}( E(x)), \mathbf{M}(\boldsymbol{\Omega}_{j}) \}
\end{array}
},
\]
where $\textsc{Sim}$ is the dot product similarity function and $E$ is a word embedding function which will be introduced in the next section.

\subsubsection{Embeddings for Word and Label Name}\label{sec:emb}

\begin{figure}[t]
	\centering
	\begin{tikzpicture}
	\draw (0,0 ) node[inner sep=0] {\includegraphics[width=1\columnwidth, trim={2.5cm 4.85cm 3.1cm 3.5cm}, clip]{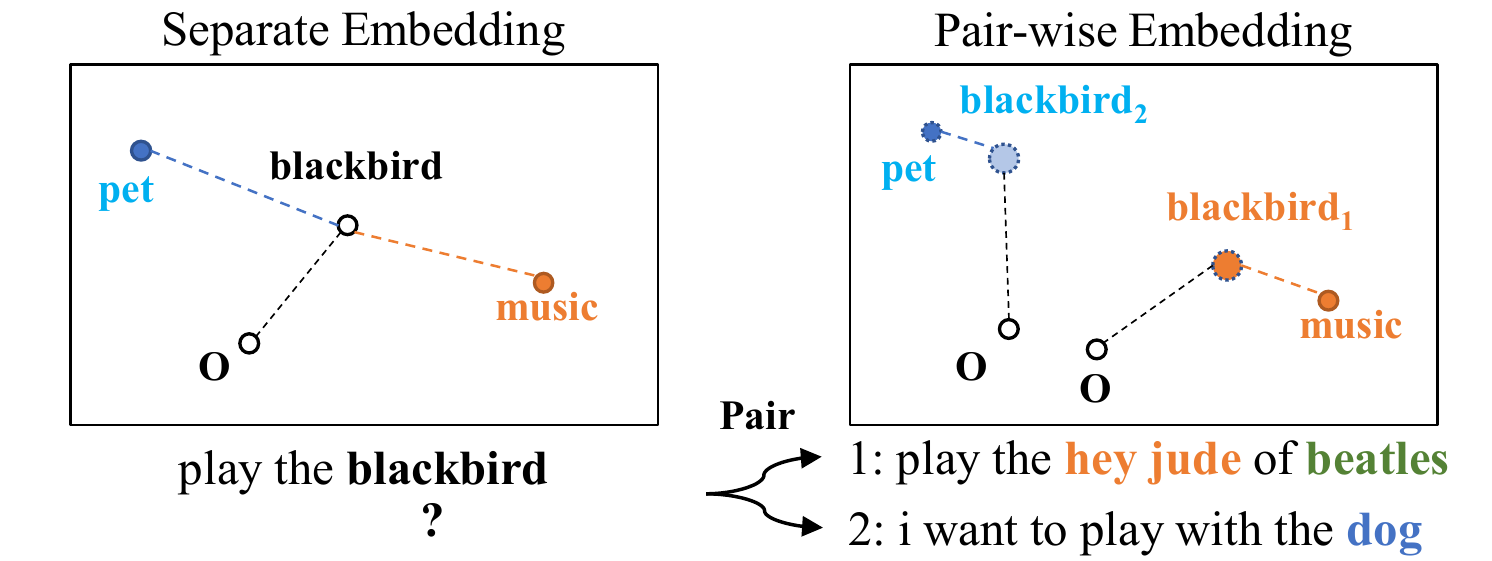}};
	\end{tikzpicture}
	\caption{\footnotesize
		An example of pair-wise embedding.
		When embedding query and support sentences separately  (left), 
		it is hard to tag \textit{blackbird} according to its similarity to labels.
		But if we embed query by pairing it with different support sentences (right), 
		the domain specific context provide \textit{blackbird} certain meanings close to \textit{pet} and \textit{song} respectively.
	}\label{fig:pair_emb}
\end{figure}

For the \textbf{word embedding function} $E$, 
we proposed a \textit{pair-wise embedding} mechanism. 
As shown in Figure \ref{fig:pair_emb}, 
a word tends to mean differently when concatenated to a different context. 
To tackle the representation challenges for similarity computation, 
we consider the special query-support setting in few-shot learning 
and embed query and support words pair-wisely.
Such pair-wise embedding can make use of domain-related context in support sentences and provide domain adaptive embeddings for the query words. 
This will further help to model the query words' similarity to domain-specific labels. 
To achieve this, 
we represent each word with self-attention over both query and support words. 
We first copy query sentence $\bm{x}$ for $N_\mathcal{S}=|\mathcal{S}|$ times, 
and pair them with all support sentences.
Then the $N_\mathcal{S}$ pairs are passed to a BERT \citep{BERT} to get $N_\mathcal{S}$ embeddings for each query word. 
We represent each word as the average of $N_\mathcal{S}$ embeddings. 
Now, representations of query words are conditioned on domain-specific context. 
We use BERT as it can naturally capture the relation between sentence pairs. 


To get label representation $\mathbf{s}$, 
we first concatenate abstract label name (e.g., \textit{begin} and \textit{inner}) and label name (e.g., \textit{weather}). 
Then, we insert a \texttt{[CLS]} token at the first position, 
and input them into a BERT. 
Finally, the representation of \texttt{[CLS]} is used as the \textbf{label semantic embedding}.

\begin{table}[t]
	\centering
	\footnotesize
	\begin{tabular}{cccccc}
		\toprule
		\multirow{2}{*}{\textbf{Domain}} & \multicolumn{2}{c}{\textbf{1-shot}} & \multicolumn{2}{c}{\textbf{5-shot}} \\
		\cmidrule(lr){2-3}
		\cmidrule(lr){4-5}
		& {\textbf{Ave. $\bm{|S|}$}} & {\textbf{Samples}} & {\textbf{Ave. $\bm{|S|}$}} & {\textbf{Samples}} \\
		\midrule
		{\textbf{We}} & 6.15 & 2,000 & 28.91 & 1,000  \\
		{\textbf{Mu}} & 7.66 & 2,000 & 34.43 & 1,000  \\
		{\textbf{Pl}} & 2.96 & 2,000 & 13.84 & 1,000  \\
		{\textbf{Bo}} & 4.34 & 2,000 & 19.83 & 1,000  \\
		{\textbf{Se}} & 4.29 & 2,000 & 19.27 & 1,000  \\
		{\textbf{Re}} & 9.41 & 2,000 & 41.58 & 1,000  \\
		{\textbf{Cr}} & 1.30 & 2,000 & 5.28 & 1,000  \\
		\bottomrule
	\end{tabular}
	\caption{\footnotesize{Overview of few-shot slot tagging data. Here, ``Ave. $|S|$" corresponds to the average support set size of each domain. And ``Sample" stands for the number of few-shot samples we build from each domain.}
	}\label{tbl:new_dataset}
	\vspace*{-3mm}
\end{table}

\section{Experiment}\label{sec:exp}
We evaluate the proposed method on {\it slot tagging} and test its generalization ability on a similar sequence labeling task: {\it name entity recognition} (NER).
Due to space limitation, we only present the detailed results for 1-shot/5-shot slot tagging, 
which transfers the learned knowledge from source domains (training) to an unseen target domain (testing) containing only a 1-shot/5-shot support set.  
The results of NER are consistent and we present them in the supplementary Appendix B.


\subsection{Settings}
\paragraph{Dataset}
For slot tagging, we exploit the \texttt{snips} dataset \citep{DBLP:journals/corr/abs-1805-10190}, 
because it contains 7 domains with different label sets and is easy to simulate the few-shot situation.
The domains are 
Weather (We), Music (Mu), PlayList (Pl), Book (Bo), Search Screen (Se), Restaurant (Re) and Creative Work (Cr).
Information about original datasets is shown in Appendix A.

To simulate the few-shot situation, 
we construct the few-shot datasets from original datasets, 
where each sample is the combination of a query data $(\bm{x^q},\bm{y^q})$ and corresponding K-shot support set $\mathcal{S}$.
Table \ref{tbl:new_dataset} shows the overview of the experiment data.

\paragraph{Few-shot Data Construction}
Different from the simple classification of single words, 
slot tagging is a structural prediction problem over the entire sentence.
So we construct support sets with sentences rather than single words under each tag. 

As a result, 
the normal \textit{N-way K-shot} few-shot definition is inapplicable for few-shot slot tagging.
We cannot guarantee that each label appears $K$ times while sampling the support sentences, 
because different slot labels randomly co-occur in one sentence. 
For example in Figure \ref{fig:intro1}, 
in the 1-shot support set, 
label \texttt{[B-weather]} occurs twice to ensure all labels appear at least once. 
So we approximately construct K-shot support set $\mathcal{S}$ following two criteria: 
(1) All labels within the domain appear at least $K$ times in $\mathcal{S}$.  
(2) At least one label will appear less than $K$ times in $\mathcal{S}$ if any $(\bm{x},\bm{y})$ pair is removed from it. 
Algorithm \ref{algorithm} shows the detail process.\footnote{
    Due to the removing step, Algorithm \ref{algorithm} has a preference for sentences with more slots. 
	So in practice, we randomly skip removing by the chance of 20\%. 
}

Here, we take the 1-shot slot tagging  as an example to illustrate the data construction procedure.
For each domain, 
we sample 100 different 1-shot support sets. 
Then, for each support set, we sample 20 unincluded utterances as queries (query set). 
Each support-query-set pair forms one \textbf{few-shot episode}. 
Eventually, we get $100$ episodes and $100\times20$ samples (1 query utterance with a support set) for each domain.


\begin{algorithm}[t]
	\caption{Minimum-including}\label{algorithm}
	\footnotesize
	\begin{algorithmic}[1]
		\Require \# of shot $K$, domain $\mathcal{D}$, label set $\mathcal{L_D}$ \\
		Initialize support set $\mathcal{S}=\left\{ \right\}$, $\text{Count}_{\ell_j} = 0 $ 
		$(\forall \ell_j \in \mathcal{L_D})$
		\\
		
		\For{$\ell$ in $\mathcal{L_D}$} {
			\While{$\text{Count}_{\ell} < k $ }
			{From $\mathcal{D} \setminus \mathcal{S}$, randomly sample a $(\bm{x}^{(i)},\bm{y}^{(i)})$ pair that $\bm{y}^{(i)}$ includes $\ell$
				
				Add $(\bm{x}^{(i)},\bm{y}^{(i)})$ to $\mathcal{S}$
				
				Update all $\text{Count}_{\ell_j}$ 
				$(\forall \ell_j \in \mathcal{L_D})$
			}
		} \\
		
		\For{each $(\bm{x}^{(i)},\bm{y}^{(i)})$ in $\mathcal{S}$}
		{   
			Remove $(\bm{x}^{(i)},\bm{y}^{(i)})$ from $\mathcal{S}$ 
			
			Update all all $\text{Count}_{\ell_j}$ 
			$(\forall \ell_j \in \mathcal{L_D})$ 
			
			\If{any $\text{Count}_{\ell_j} < $  k}
			{Put $(\bm{x}^{(i)},\bm{y}^{(i)})$ back to $\mathcal{S}$
				
				Update all $\text{Count}_{\ell_j}$ 
				$(\forall \ell_j \in \mathcal{L_D})$
			}
		}\\
		
		Return $\mathcal{S}$ 
		
	\end{algorithmic}
\end{algorithm}

\paragraph{Evaluation}
To test the robustness of our framework, 
we cross-validate the models on different domains.
Each time,
we pick one target domain for testing, one domain for development, 
and use the rest domains as source domains for training. 
So for slot tagging, 
all models are trained on 10,000 samples, and validated as well as tested on 2,000 samples respectively.

When testing model on a target domain, we evaluate F1 scores within each few-shot episode.\footnote{
	For each episode, 
	we calculate the F1 score on query samples with \texttt{conlleval} script:  \url{https://www.clips.uantwerpen.be/conll2000/chunking/conlleval.txt}
}
Then we average 100 F1 scores from all 100 episodes as the final result to counter the randomness from support-sets.
All models are evaluated on same support-query-set pairs for fairness. 

To control the nondeterministic of neural network training \citep{reimers-gurevych:2017:EMNLP2017}, 
we report the average score of 10 random seeds.

\paragraph{Hyperparameters}
We use the uncased \texttt{BERT-Base} \citep{BERT} 
to calculate contextual embeddings for all models.
We use ADAM \citep{DBLP:journals/corr/KingmaB14} to train the models with batch size 4 and a learning rate of 1e-5. 
For the CRF framework, we learn the scaling parameter $\lambda$ during training,  
which is important to get stable results. 
For L-TapNet, we set $\alpha$ as 0.5 and $\beta$ as 0.7. 
We fine-tune BERT with Gradual Unfreezing trick \cite{Howard2018UniversalLM}. 
For both proposed and baseline models, we take early stop in training and fine-tuning when there is no loss decay withing a fixed number of steps.

\begin{table*}[t]
	\centering
	\footnotesize
	\begin{tabular}{lrrrrrrrrr}\toprule
		\multirow{2}{*}{\textbf{Model}} &
		\multicolumn{7}{c}{\textbf{1-shot Slot Tagging}} & \\
		\cmidrule(lr){2-8}
		& \multicolumn{1}{c}{\textbf{We}} & \multicolumn{1}{c}{\textbf{Mu}} & \multicolumn{1}{c}{\textbf{Pl}} & \multicolumn{1}{c}{\textbf{Bo}} & \multicolumn{1}{c}{\textbf{Se}} & \multicolumn{1}{c}{\textbf{Re}} & \multicolumn{1}{c}{\textbf{Cr}} & \multicolumn{1}{c}{\textbf{Ave.}} \\
		\midrule
		Bi-LSMT & { 10.36 } & { 17.13 } & { 17.52 } & { 53.84 } & { 18.44 } & { 22.56 } & { 8.64 } & { 21.21 }\\
		SimBERT & { 36.10 } & { 37.08 } & { 35.11 } & { 68.09 } & { 41.61 } & { 42.82 } & { 23.91 } & { 40.67 }\\
		TransferBERT & { 55.82 } & { 38.01 } & { 45.65 } & { 31.63 } & { 21.96 } & { 41.79 } & { 38.53 } & { 39.06 }\\
		MN & { 21.74 } & { 10.68 } & { 39.71 } & { 58.15 } & { 24.21 } & { 32.88 } & { \textbf{69.66} } & { 36.72 }\\
		WPZ & { 4.53 } & { 7.43 } & { 14.43 } & { 39.15 } & { 11.69 } & { 7.78 } & { 10.09 } & { 13.59 }\\
		WPZ+GloVe & { 17.92 } & { 22.37 } & { 19.90 } & { 42.61 } & { 22.30 } & { 22.79 } & { 16.75 } & { 23.52 }\\
		WPZ+BERT & { 46.72 } & { 40.07 } & { 50.78 } & { 68.73 } & { 60.81 } & { 55.58 } & { 67.67 } & { 55.77 }\\
		
		\midrule
		TapNet & { 51.12 } & { 40.65 } & { 48.41 } & { 77.50 } & { 49.77 } & { 54.79 } & { 61.39 } & { 54.80 }\\
		TapNet+CDT & { 66.30 } & { 55.93 } & { 57.55 } & { 83.32 } & { 64.45 } & { 65.65 } & { 67.91 } & { 65.87 }\\
		L-WPZ+CDT & { 71.23 } & { 47.38 } & { 59.57 } & { 81.98 } & { 69.83 } & { 66.52 } & { 62.84 } & { 65.62 }\\
		L-TapNet+CDT & { \textbf{71.53} } & { \textbf{60.56} } & { \textbf{66.27} } & { \textbf{84.54} } & { \textbf{76.27} } & { \textbf{70.79} } & { 62.89 } & { \textbf{70.41} }\\

		\bottomrule
	\end{tabular}
	\caption{ \footnotesize  
		F1 scores on 1-shot slot tagging. 
        \texttt{+CDT} denotes \texttt{collapsed dependency transfer}.
		Score below mid-line are from our methods, which achieve the best performance. 
		Ave. shows the averaged scores. Results with standard deviations is showed in Appendix \ref{sec:std}.
	}\label{tbl:1shot}
\end{table*}

\begin{table*}[t]
	\centering
	\footnotesize
	\begin{tabular}{lrrrrrrrrr}\toprule
		\multirow{2}{*}{\textbf{Model}} &
		\multicolumn{7}{c}{\textbf{5-shots Slot Tagging}} & \\
		\cmidrule(lr){2-8}
		& \multicolumn{1}{c}{\textbf{We}} & \multicolumn{1}{c}{\textbf{Mu}} & \multicolumn{1}{c}{\textbf{Pl}} & \multicolumn{1}{c}{\textbf{Bo}} & \multicolumn{1}{c}{\textbf{Se}} & \multicolumn{1}{c}{\textbf{Re}} & \multicolumn{1}{c}{\textbf{Cr}} & \multicolumn{1}{c}{\textbf{Ave.}} \\

		\midrule
		Bi-LSMT & { 25.17 } & { 39.80 } & { 46.13 } & { 74.60 } & { 53.47 } & { 40.35 } & { 25.10 } & { 43.52 }\\
		SimBERT & { 53.46 } & { 54.13 } & { 42.81 } & { 75.54 } & { 57.10 } & { 55.30 } & { 32.38 } & { 52.96 }\\
		TransferBERT & { 59.41 } & { 42.00 } & { 46.07 } & { 20.74 } & { 28.20 } & { 67.75 } & { 58.61 } & { 46.11 }\\
		MN & { 36.67 } & { 33.67 } & { 52.60 } & { 69.09 } & { 38.42 } & { 33.28 } & { 72.10 } & { 47.98 }\\
		WPZ & { 9.54 } & { 14.23 } & { 18.12 } & { 44.65 } & { 18.98 } & { 12.03 } & { 14.05 } & { 18.80 }\\
		WPZ+GloVe & { 26.61 } & { 34.25 } & { 22.11 } & { 50.55 } & { 28.53 } & { 34.16 } & { 23.69 } & { 31.41 }\\
		WPZ+BERT & { 67.82 } & { 55.99 } & { 46.02 } & { 72.17 } & { 73.59 } & { 60.18 } & { 66.89 } & { 63.24 }\\
		
		\midrule
		
		TapNet & { 53.03 } & { 49.80 } & { 54.90 } & { 83.36 } & { 63.07 } & { 59.84 } & { 67.02 } & { 61.57 }\\
		TapNet+CDT & { 66.48 } & { 66.36 } & { 68.23 } & { \textbf{85.76} } & { 73.60 } & { 64.20 } & { 68.47 } & { 70.44 }\\
		L-WPZ+CDT & { \textbf{74.68} } & { 56.73 } & { 52.20 } & { 78.79 } & { 80.61 } & { 69.59 } & { 67.46 } & { 68.58 }\\
		L-TapNet+CDT & { 71.64 } & { \textbf{67.16} } & { \textbf{75.88} } & { 84.38 } & { \textbf{82.58} } & { \textbf{70.05} } & { \textbf{73.41} } & { \textbf{75.01} }\\
		\bottomrule
	\end{tabular}
	\caption{ \footnotesize 
		F1 score results on 5-shots slot tagging. 
		Our methods achieve the best performance. Results with standard deviations is showed in Appendix \ref{sec:std}.
	}\label{tbl:5shot}
	\vspace*{-4mm}
\end{table*}

\subsection{Baselines}

\paragraph{Bi-LSTM} is a bidirectional LSTM \cite{birnn} with GloVe \citep{GloVe} embedding for slot tagging.
It is trained on the support set and tested on the query samples.

\paragraph{SimBERT} is a model that predicts labels according to cosine similarity of word embedding of non-fine-tuned BERT. 
For each word $x_j$, SimBERT finds its most similar word $x_k'$ in support set, and the label of $x_j$ is predicted to be the label of $x_k'$. 

\paragraph{TransferBERT} is a domain transfer model with the NER setting of BERT \citep{BERT}. 
We pretrain the it on source domains and select the best model on the same dev set of our model. 
We deal with label mismatch by only transferring bottleneck feature. 
Before testing, we fine-tune it on target domain support set. Learning rate is set as 1e-5 in training and fine-tuning. 

\paragraph{WarmProtoZero (WPZ)} \citep{baseline} is a few-shot sequence labeling model 
that regards sequence labeling as classification of every single word. 
It pre-trains a prototypical network \citep{prototypical} on source domains, 
and utilize it to do word-level classification on target domains without training. 
\citet{baseline} use randomly initialized word embeddings.
To eliminate the influence of different embedding methods, 
we further implement WPZ with the pre-trained embedding of GloVe \citep{GloVe} and BERT.  

\paragraph{Matching Network (MN)} is similar to WPZ. 
The only difference is that we employ the matching network \cite{matching} with BERT embedding for classification.

\subsection{Main Results}
\paragraph{Results of 1-shot Setting}
Table \ref{tbl:1shot} shows the 1-shot slot tagging results. 
Each column respectively shows the F1 scores of taking a certain domain as target domain (test) and use others as source domain (train \& dev).
As shown in the tables, 
our L-TapNet+CDT 
achieves the best performance. 
It outperforms the strongest few-shot learning baseline WPZ+BERT by average F1 scores of 14.64. 


Our model significantly outperforms Bi-LSTM and TransferBERT, 
indicating that the number of labeled data under the few-shot setting is too scarce for 
both conventional machine learning and transfer learning models. 
Moreover, the performance of SimBERT demonstrates the superiority of metric-based methods 
over conventional machine learning models in the few-shot setting.

The original WarmProtoZero (WPZ) model suffers from the weak representation ability of its word embeddings. 
When we enhance it with GloVe and BERT word embeddings, 
its performance improves significantly.
This shows the importance of embedding in the few-shot setting. 
Matching Network (MN) performs poorly in both settings.
This is largely due to the fact that MN pays attention to all support word equally, which makes it vulnerable to the unbalanced amount of O-labels. 

More specifically, those models that are fine-tuned on support set, such as Bi-LSTM and TransferBERT, tend to predict tags randomly. 
Those systems can only handle the cases that are easy to generalize from support examples, such as tags for proper noun tokens (e.g. city name and time). 
This shows that fine-tuning on extremely limited examples leads to poor generalization ability and undertrained classifier. 
And for those metric based methods, such as WPZ and MN, label prediction is much more reasonable. 
However, these models are easy to be confused by similar labels, such as \textit{current\_location} and \textit{geographic\_poi}. 
It indicates the necessity of well-separated label representations. 
Also illegal label transitions are very common, which can be well tackled by the proposed collapsed dependency transfer.


To eliminate unfair comparisons caused by additional information in label names,
we propose the \textbf{L-WPZ+CDT} 
by enhancing the WarmProtoZero (WPZ) model
with label name representation same to L-TapNet and incorporating it into the proposed CRF framework.  
It combines label name embedding and prototype as each label representation.
Its improvements over WPZ mainly come from 
label semantics, collapsed dependency transfer and pair-wise embedding. 
L-TapNet+CDT outperforms L-WPZ+CDT by 4.79 F1 scores demonstrating the effectiveness of embedding projection. 
When compared with TapNet+CDT, 
L-TapNet+CDT achieves an improvement of 4.54 F-score on average, which shows that considering label semantics and prototype helps improve emission score calculation. 


\paragraph{Results of 5-shots Setting}
Table \ref{tbl:5shot} shows the results of 5-shots experiments,
which verify the proposed model's generalization ability in more shots situations. 
The results are consistent with 1-shot setting in general trending. 

\subsection{Analysis}
\paragraph{Ablation Test}
To get further an understanding of each component in our method (L-TapNet+CDT), 
we conduct ablation analysis on both 1-shot and 5-shots setting in Table \ref{tbl:ablation}.
Each component of our method is removed respectively, 
including: 
\textit{collapsed dependency transfer}, \textit{pair-wise embedding}, \textit{label semantic}, and \textit{prototype reference}.

When collapsed dependency transfer is removed, 
we directly predict labels with emission score and huge F1 score drops are witnessed in all settings. 
This ablation demonstrates a great necessity for considering label dependency. 

For our method without pair-wise embedding, 
we represent query and support sentences independently.
We address the drop to the fact that 
support sentences can provide domain-related context, 
and pair-wise embedding can leverage such context and provide domain-adaptive representation for words in query sentences. 
This helps a lot when computing a word's similarity to domain-specific labels. 

When we remove the label-semantic from L-TapNet, 
the model degenerates into TapNet+CDT enhanced with prototype in emission score.
The drops in results show that considering label name can provide better label representation 
and help to model word-label similarity.
Further, we also tried to remove the inner and beginning words in label representation and observe a 0.97 F1-score drop on 1-shot SNIPS. 
It shows that distinguishing B-I labels in label semantics can help tagging.

And if we calculate emission score without the prototype reference, 
the model loses more performance in 5-shots setting. 
This meets the intuition that prototype allows model to benefit more from the increase of support shots, 
as prototypes are directly derived from the support set.



\begin{table}[t]
	\centering
	\footnotesize
	\begin{tabular}{lrr}
		\toprule
		{\textbf{Model}} & \multicolumn{1}{c}{1-shot} & \multicolumn{1}{c}{5-shots}  \\
		\midrule
		Ours & { 70.41 } & { 75.01 } \\
		~ - dependency transfer & { -10.01 } & { -8.08 } \\
		~ - pair-wise embedding & { -8.29 } & { -7.74 }  \\
		~ - label semantic & { -9.57 } & { -4.87 } \\
		~ - prototype reference & { -1.73 } &  { -3.33 } \\
		\bottomrule
	\end{tabular}
	\caption{
		\footnotesize
		Ablation test over different components on slot tagging task. 
		Results are averaged F1-score of all domains.
	}\label{tbl:ablation}
\end{table}

\begin{table}[t]
	\centering
	\footnotesize
	\begin{tabular}{lSS}
		\toprule
		\textbf{Model} & {1-shot} & {5-shots} \\
		\midrule
		L-TapNet & 60.40 & 66.93 \\
		L-TapNet+Rule & 65.30 & 69.64 \\
		L-TapNet+CDT & {\textbf{70.41}} & {\textbf{75.01}} \\
		
		\bottomrule
	\end{tabular}
	\caption{
		\footnotesize
		Comparison between transition rules and collapsed dependency transfer (CDT). 
	}\label{tbl:rule}
\end{table}

\begin{table}[t]
	\centering
	\footnotesize
	\begin{tabular}{llScc}
		\toprule
		\multicolumn{2}{c}{\textbf{Bi-gram Type}} & \textbf{Proportion} & \textbf{L-TapNet} & \textbf{+CDT}\\
		\cmidrule(lr){1-2}
		\cmidrule(lr){3-5}
		\multirow{4}{*}{\makecell{Border}} 
		& O-O  & {28.5\%}  & 82.7\% & \textbf{83.7\%} \\ 
		& O-B  & {24.5\%}  & 78.3\% & \textbf{81.5\%} \\
		& B-O  & {8.2\%}   & 72.4\% & \textbf{74.8\%} \\
		& I-O  & {5.8\%}   & 76.7\% &  \textbf{81.7\%} \\
		& I-B/B-B & {7.8\%} &  65.0\% &  \textbf{72.5\%} \\
		\midrule
		\multirow{2}{*}{\makecell{Inner}} 
		& B-I  & {13.3\%}  & 78.5\% & \textbf{83.6\%} \\
		& I-I  & {12.1\%}  & 77.8\% & \textbf{82.7\%} \\
		\bottomrule
	\end{tabular}
	\caption{\footnotesize
		Accuracy analysis of label prediction on 1-shot slot tagging. 
		The table shows accuracy and proportion of different bi-gram types in dataset.
	}\label{tbl:error_rd}
	\vspace*{-4mm}
\end{table}

\paragraph{Analysis of Collapsed Dependency Transfer}
While collapsed dependency transfer (CDT) brings significant improvements, 
two natural questions arise:  whether CDT just learns simple transition rules and why it works. 

To answer the first question, 
we replace CDT with transition rules in Table \ref{tbl:rule},\footnote{
	Transition Rule:
	We greedily predict the label for each word and block the result that conflicts with previous label. 
}
which shows CDT can bring more improvements than transition rules. 

To have a deeper insight into the effectiveness of CDT, 
we conduct an accuracy analysis of it. 
We assess the label predicting accuracy of different types of label bi-grams. 
The result is shown in Table \ref{tbl:error_rd}. 
We further summarize the bi-grams into 2 categories: 
\textbf{Border} includes the bi-grams across the border of a slot span; 
\textbf{Inner} is the bi-grams within a slot span.
We argue that improvements of \textbf{Inner} show successful reduction of illegal label transition from CDT. 
Interestingly, we observe that CDT also brings improvements by correctly predict the first and last token of a slot span.
The results of \textbf{Border} verified our observation that CDT may helps to decide the boundaries of slot spans more accurately, 
which is hard to achieve by adding transition rules.

\vspace*{-1mm}
\section{Related Works}	
\vspace*{-1mm}
Traditional few-shot learning methods depend highly on hand-crafted features \cite{fei2006knowledge,fink2005object}. 
Classical methods primarily focus on metric learning \cite{prototypical,matching}, 
which classifies an item according to its similarity to each class's representation. 
Recent efforts \cite{lu2018zero,schwartz2019baby} propose to leverage the semantics of class name to enhance class representation. 
However, different from us, these methods focus on image classification 
where effects of name semantic are implicit and label dependency is not required.

Few-shot learning in natural language processing has been explored for classification tasks, including  
text classification \cite{textSun2019hierarchical,textGeng2019induction,yan2018few,yu2018diverse}, 
entity relation classification \cite{relationLv2019adapting,relationGao2019neural,relationYeL19}, and dialog act prediction \cite{policyVlasov2018few}. 
However, few-shot learning for slot tagging is less investigated. 
\citet{slotLuo2018marrying} investigated few-shot slot tagging using additional regular expressions, 
which is not comparable to our model due to the usage of regular expressions. 
\citet{baseline} explored few-shot named entity recognition with the Prototypical Network, 
which has a similar setting to us.
Compared to it, our model achieves better performance by 
considering both label dependency transferring and label name semantics. 
Zero-shot slot tagging methods \cite{zeroBapna2017towards,zeroLee2019zero,zeroSlotShah2019robust} share a similar idea to us in using label name semantics, 
but has a different setting as few-shot methods are additionally supported by a few labeled sentences.  
\citet{chen2016zero} investigate using label name in intent detection. 
In addition to learning directly from limited example, another research line of solving data scarcity problem in NLP is data augmentation
\cite{fader-zettlemoyer-etzioni:2013:ACL2013,zhang2015character,liu2017generating}. 
For data augmentation of slot tagging, 
sentence generation based methods are explored to create additional labeled samples \cite{hou2018coling,shin2019vae,yoo2019joint}. 

\vspace*{-1mm}
\section{Conclusion}\label{sec:con}
\vspace*{-1mm}
In this paper, 
we propose a few-shot CRF model for slot tagging of task-oriented dialogue. 
To compute transition score under few-shot setting, 
we propose the collapsed dependency transfer mechanism, 
which transfers the prior knowledge of the label dependencies across domains with different label sets.
And we propose L-TapNet to calculate emission score, which improves label representation with label name semantics. 
Experiment results validate that both the collapsed dependency transfer and L-TapNet can improve the tagging accuracy. 


\section*{Acknowledgments}
We sincerely thank Ning Wang and Jiafeng Mao for the help on both paper and experiments. 
We are grateful for the helpful comments and suggestions from the anonymous reviewers.
This work was supported by the National Natural Science Foundation of China (NSFC) via grant 61976072, 61632011 and 61772153.

\bibliography{acl2020}
\bibliographystyle{acl_natbib}

\appendix

\section*{Appendix}

\section{Detail of Dataset}\label{apx:old_data}
Table \ref{tbl:dataset_old} shows the statistics of the original dataset used to construct few-shot experiment data. 
\begin{table}[h!]
	\centering
	\scriptsize
	\begin{tabular}{SlllSS} \toprule
		{\textbf{Task}}  & {\textbf{Dataset}}  & {\textbf{Domain}}  & {\textbf{\# Sent}} & {\textbf{\# Labels}}  \\ 
		\midrule
		{\multirow{7}{*}{\makecell{Slot\\Tagging}}}	& {\multirow{7}{*}{Snips}}	& {We}& {2,100}  & {10}   \\		
		& & {Mu}& {2,100}  & {10}  \\
		& & {Pl} & {2,042} & {6}  \\		
		& & {Bo} & {2056}  & {8} \\
		& & {Se} & {2,059}  & {8}  \\
		& & {Re} & {2,073}  & {15}   \\		
		& & {Cr} & {2,054}  & {3}  \\
		\midrule
		{\multirow{4}{*}{NER}} & {CoNLL}	& {News}  & {20679}  & {5}   \\
		& {GUM}	& {WiKi} & {3,493}  & {12}    \\
		& {WNUT} & {Social} & {5,657} & {7}    \\
		& {OntoNotes}  & {Mixed}      & {159,615}   & {19}       \\
		\bottomrule
	\end{tabular}
	\caption{\footnotesize
		Statistic of Original Dataset}\label{tbl:dataset_old}
	\vspace*{-3mm}
\end{table}

\section{Few-shot experiments for Name entity recognition}\label{apx:ner}
Name entity recognition (NER) 
that identify pre-defined name entities, such as the person names, organizations and locations, 
can be modeled as a slot tagging task. 
Also, the data scarcity problem for a new domain exists in the NER task.
For the above reasons, 
we conduct few-shot NER experiments to test our model's generation ability. 

\begin{table}[h!]
	\centering
	\scriptsize
	\begin{tabular}{cccccc}
		\toprule
		\multirow{2}{*}{\textbf{Domain}} & \multicolumn{2}{c}{\textbf{1-shot}} & \multicolumn{2}{c}{\textbf{5-shots}} \\
		\cmidrule(lr){2-3}
		\cmidrule(lr){4-5}
		& {\textbf{Ave. $\bm{|S|}$}} & {\textbf{Samples}} & {\textbf{Ave. $\bm{|S|}$}} & {\textbf{Samples}} \\
		\midrule
		{\textbf{News}} & 3.38 & 4,000 & 15.58 & 1,000  \\
		{\textbf{Wiki}} & 6.50 & 4,000 & 27.81 & 1,000  \\
		{\textbf{Social}} & 5.48 & 4,000 & 28.66 & 1,000  \\
		{\textbf{Mixed}} & 14.38 & 2,000 & 62.28 & 1,000  \\
		\bottomrule
	\end{tabular}
	\caption{\footnotesize{Overview of few-shot data for NER experiments. Here, ``Ave. $|S|$" corresponds to the average support set size of each domain. And ``Sample" stands for the number of few-shot samples we build from each domain.}
	}\label{tbl:new_dataset_ner}
	\vspace*{-3mm}
\end{table}

\begin{table*}[h!]
	\centering
	\scriptsize
	\begin{tabular}{p{1.7cm}rrrrrrr}\toprule
		\multirow{2}{*}{\textbf{Model}} &
		\multicolumn{4}{c}{\textbf{1-shot Named Entity Recognition}}  &	\\
		\cmidrule(lr){2-5}
		
		& \multicolumn{1}{c}{\textbf{News}} & \multicolumn{1}{c}{\textbf{Wiki}} & \multicolumn{1}{c}{\textbf{Social}} & \multicolumn{1}{c}{\textbf{Mixed}} & \multicolumn{1}{c}{\textbf{Ave.}} \\
		\midrule
		Bi-LSMT & { 2.57  \tiny{$\pm 0.14$} } & { 3.29  \tiny{$\pm 0.19$} } & { 0.67  \tiny{$\pm 0.07$} } & { 2.11  \tiny{$\pm 0.15$} } & { 2.16  \tiny{$\pm 0.14$} }\\
		SimBERT & { 19.22  \tiny{$\pm 0.00$} } & { 6.91  \tiny{$\pm 0.00$} } & { 5.18  \tiny{$\pm 0.00$} } & { 13.99  \tiny{$\pm 0.00$} } & { 11.32  \tiny{$\pm 0.00$} }\\
		TransferBERT & { 4.75  \tiny{$\pm 1.42$} } & { 0.57  \tiny{$\pm 0.32$} } & { 2.71  \tiny{$\pm 0.72$} } & { 3.46  \tiny{$\pm 0.54$} } & { 2.87  \tiny{$\pm 0.75$} }\\
		MN & { 19.50  \tiny{$\pm 0.35$} } & { 4.73  \tiny{$\pm 0.16$} } & { 17.23  \tiny{$\pm 2.75$} } & { 15.06  \tiny{$\pm 1.61$} } & { 14.13  \tiny{$\pm 1.22$} }\\
		WPZ & { 3.64  \tiny{$\pm 0.08$} } & { 2.00  \tiny{$\pm 0.02$} } & { 0.92  \tiny{$\pm 0.04$} } & { 0.66  \tiny{$\pm 0.03$} } & { 1.80  \tiny{$\pm 0.04$} }\\
		WPZ+GloVe & { 9.40  \tiny{$\pm 0.06$} } & { 3.23  \tiny{$\pm 0.01$} } & { 2.29  \tiny{$\pm 0.02$} } & { 2.56  \tiny{$\pm 0.01$} } & { 4.37  \tiny{$\pm 0.03$} }\\
		WPZ+BERT & { 32.49  \tiny{$\pm 2.01$} } & { 3.89  \tiny{$\pm 0.24$} } & { 10.68  \tiny{$\pm 1.40$} } & { 6.67  \tiny{$\pm 0.46$} } & { 13.43  \tiny{$\pm 1.03$} }\\
		\midrule
		L-TapNet+CDT & { \textbf{44.30}  \tiny{$\pm 3.15$} } & { \textbf{12.04}  \tiny{$\pm 0.65$} } & { \textbf{20.80}  \tiny{$\pm 1.06$} } & { \textbf{15.17}  \tiny{$\pm 1.25$} } & { \textbf{23.08}  \tiny{$\pm 1.53$} }\\
		\bottomrule
	\end{tabular}
	\caption{ \footnotesize  
			F1 scores on 1-shot name entity recognition. 
			\texttt{CDT} denotes \texttt{collapsed dependency transfer}. 
			Scores below mid-line are from our models, which achieve the best performance. 
			Ave. shows the averaged scores. 
		}\label{tbl:1shotner}
\end{table*}

\begin{table*}[h!]
	\centering
	\scriptsize
	\begin{tabular}{p{1.7cm}rrrrrrr}\toprule
		\multirow{2}{*}{\textbf{Model}} &
		\multicolumn{4}{c}{\textbf{5-shots Named Entity Recognition}}  & 	\\
		\cmidrule(lr){2-5}
		& \multicolumn{1}{c}{\textbf{News}} & \multicolumn{1}{c}{\textbf{Wiki}} & \multicolumn{1}{c}{\textbf{Social}} & \multicolumn{1}{c}{\textbf{Mixed}} & \multicolumn{1}{c}{\textbf{Ave.}} \\
		\midrule
		Bi-LSMT & { 6.81  \tiny{$\pm 0.40$} } & { 8.40  \tiny{$\pm 0.16$} } & { 1.06  \tiny{$\pm 0.16$} } & { 13.17  \tiny{$\pm 0.17$} } & { 7.36  \tiny{$\pm 0.22$} }\\
		SimBERT & { 32.01  \tiny{$\pm 0.00$} } & { 10.63  \tiny{$\pm 0.00$} } & { 8.20  \tiny{$\pm 0.00$} } & { 21.14  \tiny{$\pm 0.00$} } & { 18.00  \tiny{$\pm 0.00$} }\\
		TransferBERT & { 15.36  \tiny{$\pm 2.81$} } & { 3.62  \tiny{$\pm 0.57$} } & { 11.08  \tiny{$\pm 0.57$} } & { \textbf{35.49}  \tiny{$\pm 7.60$} } & { 16.39  \tiny{$\pm 2.89$} }\\
		MN & { 19.85  \tiny{$\pm 0.74$} } & { 5.58  \tiny{$\pm 0.23$} } & { 6.61  \tiny{$\pm 1.75$} } & { 8.08  \tiny{$\pm 0.47$} } & { 10.03  \tiny{$\pm 0.80$} }\\
		WPZ & { 4.09  \tiny{$\pm 0.16$} } & { 3.19  \tiny{$\pm 0.13$} } & { 0.86  \tiny{$\pm 0.23$} } & { 0.93  \tiny{$\pm 0.14$} } & { 2.27  \tiny{$\pm 0.17$} }\\
		WPZ+GloVe & { 16.94  \tiny{$\pm 0.10$} } & { 5.33  \tiny{$\pm 0.07$} } & { 5.53  \tiny{$\pm 0.12$} } & { 3.54  \tiny{$\pm 0.03$} } & { 7.83  \tiny{$\pm 0.08$} }\\
		WPZ+BERT & { \textbf{50.06}  \tiny{$\pm 1.57$} } & { 9.54  \tiny{$\pm 0.44$} } & { 17.26  \tiny{$\pm 2.65$} } & { 13.59  \tiny{$\pm 1.61$} } & { 22.61  \tiny{$\pm 1.57$} }\\
		\midrule
		L-TapNet+CDT & { 45.35  \tiny{$\pm 2.67$} } & { \textbf{11.65}  \tiny{$\pm 2.34$} } & { \textbf{23.30}  \tiny{$\pm 2.80$} } & { 20.95  \tiny{$\pm 2.81$} } & { \textbf{25.31}  \tiny{$\pm 2.65$} }\\
		\bottomrule
	\end{tabular}
	\caption{ \footnotesize.  
		F1 score results on 5-shots name entity recognition. 
		Our methods achieve the best performance.
	}\label{tbl:5shotner}
	\vspace*{-4mm}
\end{table*}

\paragraph{Experiment Data for Few-shot NER}
For named entity recognition, we utilize 4 different datasets: 
\texttt{CoNLL-2003} \citep{sang2003introduction}, \texttt{GUM} \citep{zeldes2017gum}, \texttt{WNUT-2017} \citep{derczynski2017results} and \texttt{Ontonotes} \citep{pradhan2013towards}, 
each of which contains data from only 1 domain. 
The 4 domains are News, Wiki, Social and Mixed. 
Detail of the original data set is showed in Table \ref{tbl:dataset_old} and statistic of constructed few-shot data is showed in Table \ref{tbl:new_dataset_ner}.

\paragraph{1-shot and 5-shots Results for NER}
Table \ref{tbl:1shotner} and Table \ref{tbl:5shotner} respectively show the 1-shot and 5-shots name entity recognition results. Our best model outperforms all baseline in both settings. 

The trend of results is consistent with slot-tagging results. 
But the overall score is much lower than slot-tagging results. 
this is because NER domains are from different datasets and the domain gap is much larger. 

Our improvements on 5-shots is narrowed in margin. This is because NER domains have different genres and vocabulary. 
So compared to SNIPS, it is harder to transfer knowledge but benefits more to rely on domain-specific support examples. 
This trend is even more pronounced with more shots. 
In 5-shots setting, the strongest baseline WPZ benefits more from the increased shots because it only uses support set for prediction.
But the benefit of more shots is weaker for our model because it uses more prior knowledge.

\paragraph{Ablation Analysis on NER}

\begin{table}[t]
	\centering
	\footnotesize
	\begin{tabular}{lSS}
		\toprule
		{\textbf{Model}} & {1-shot} & {5-shots}  \\
		\midrule
		Ours & 22.19 & 24.12 \\
		~ - dependency transfer & { -4.55 } & { -4.83 } \\
		~ - label semantic & { -6.93 } & { -1.46 } \\
		\bottomrule
	\end{tabular}
	\caption{
		\footnotesize
		Ablation test over different components on NER task. 
		Results are averaged F1-score of all domains.
	}\label{tbl:ner_ablation}
	\vspace*{-4mm}
\end{table}

We investigate effectiveness of collapsed dependency transfer and label semantic on the NER task. 
We perform ablations on two proposed components and observe performance drops on both 1-shot and 5-shots settings, 
which demonstrate the generalization ability of proposed two mechanism.  

\section{Analysis of Projection Space Dimensionality}

\begin{figure}[t]
	\centering
	\begin{tikzpicture}
	\draw (0,0 ) node[inner sep=0] {\includegraphics[width=0.85 \columnwidth, trim={0cm 0.8cm 0cm 0.68cm}, clip]{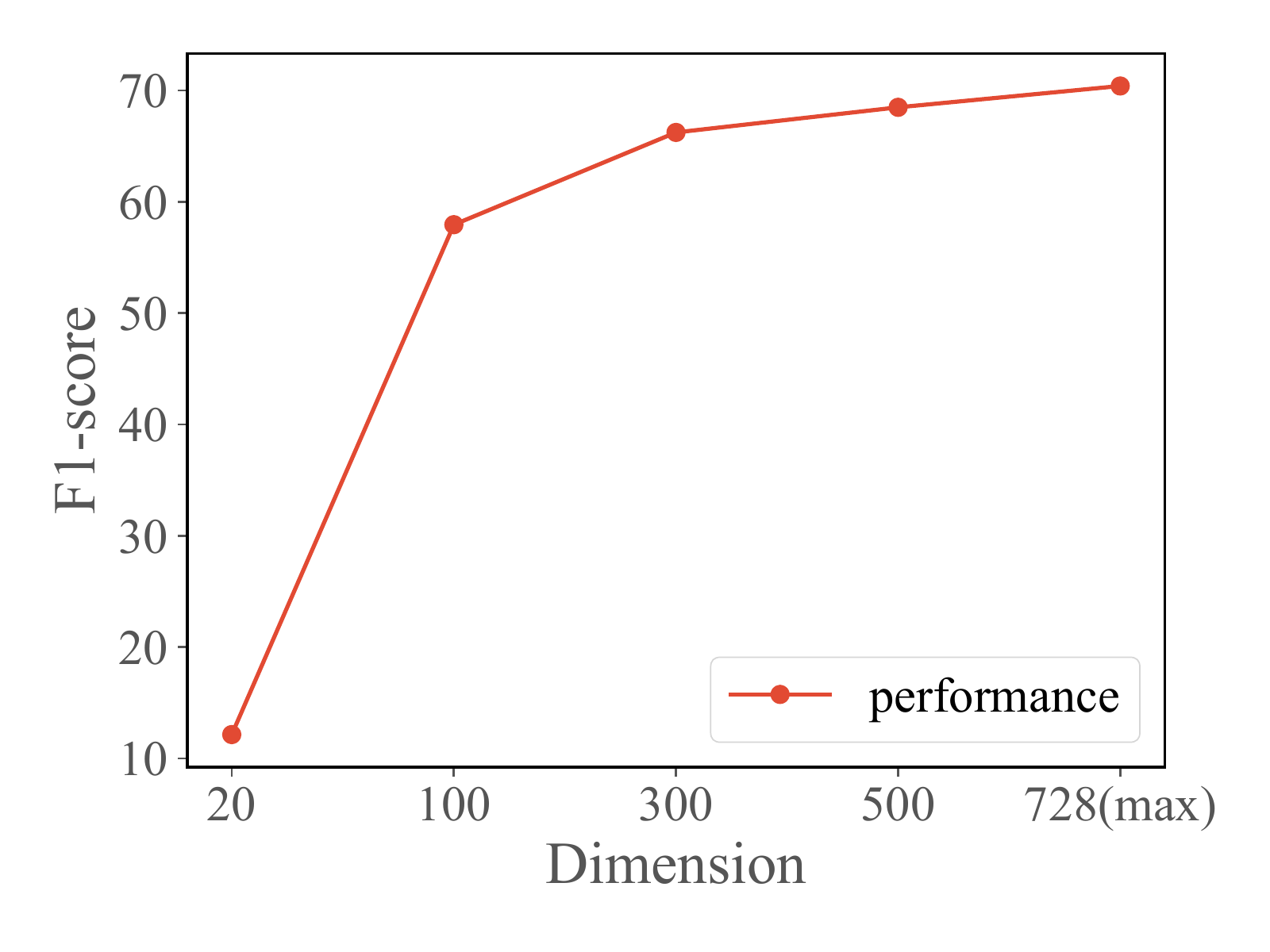}};
	\end{tikzpicture}
	\caption{\footnotesize
	 Impacts of projection space dimensionality. 
	 }\label{fig:project_dim}
 	\vspace*{-4mm}
\end{figure}

Fig \ref{fig:project_dim} shows the performance on 1-shot Snips when using different projected-space dimensions in L-TapNet.
As shown in the trend in the figure, 
the performance of the model becomes better as the dimension of the mapping space increases and gradually stabilizes.
This shows the possibility of reducing the dimension without losing too much performance \cite{yoon2019tapnet}.

\section{Slot Tagging Result with Standard Deviations}\label{sec:std}
Table \ref{tbl:1shot_std} and \ref{tbl:5shot_std} show the complete results with standard deviations for slot tagging task. 

\begin{table*}[t]
	\centering
	\scriptsize
	\resizebox{\linewidth}{!}{
	\begin{tabular}{lrrrrrrrrr}\toprule
		\multirow{2}{*}{\textbf{Model}} &
		\multicolumn{7}{c}{\textbf{1-shot Slot Tagging}} & \\
		\cmidrule(lr){2-8}
		& \multicolumn{1}{c}{\textbf{We}} & \multicolumn{1}{c}{\textbf{Mu}} & \multicolumn{1}{c}{\textbf{Pl}} & \multicolumn{1}{c}{\textbf{Bo}} & \multicolumn{1}{c}{\textbf{Se}} & \multicolumn{1}{c}{\textbf{Re}} & \multicolumn{1}{c}{\textbf{Cr}} & \multicolumn{1}{c}{\textbf{Ave.}} \\
		\midrule
		Bi-LSMT & { 10.36\tiny{$\pm 0.36$} } & { 17.13\tiny{$\pm 0.61$} } & { 17.52\tiny{$\pm 0.76$} } & { 53.84\tiny{$\pm 0.57$} } & { 18.44\tiny{$\pm 0.44$} } & { 22.56\tiny{$\pm 0.10$} } & { 8.64\tiny{$\pm 0.41$} } & { 21.21\tiny{$\pm 0.46$} }\\
		SimBERT & { 36.10\tiny{$\pm 0.00$} } & { 37.08\tiny{$\pm 0.00$} } & { 35.11\tiny{$\pm 0.00$} } & { 68.09\tiny{$\pm 0.00$} } & { 41.61\tiny{$\pm 0.00$} } & { 42.82\tiny{$\pm 0.00$} } & { 23.91\tiny{$\pm 0.00$} } & { 40.67\tiny{$\pm 0.00$} }\\
		TransferBERT & { 55.82\tiny{$\pm 2.75$} } & { 38.01\tiny{$\pm 1.74$} } & { 45.65\tiny{$\pm 2.02$} } & { 31.63\tiny{$\pm 5.32$} } & { 21.96\tiny{$\pm 3.98$} } & { 41.79\tiny{$\pm 3.81$} } & { 38.53\tiny{$\pm 7.42$} } & { 39.06\tiny{$\pm 3.86$} }\\
		MN & { 21.74\tiny{$\pm 4.60$} } & { 10.68\tiny{$\pm 1.07$} } & { 39.71\tiny{$\pm 1.81$} } & { 58.15\tiny{$\pm 0.68$} } & { 24.21\tiny{$\pm 1.20$} } & { 32.88\tiny{$\pm 0.64$} } & { \textbf{69.66}\tiny{$\pm 1.68$} } & { 36.72\tiny{$\pm 1.67$} }\\
		WPZ & { 4.53\tiny{$\pm 0.18$} } & { 7.43\tiny{$\pm 0.31$} } & { 14.43\tiny{$\pm 0.73$} } & { 39.15\tiny{$\pm 1.10$} } & { 11.69\tiny{$\pm 0.16$} } & { 7.78\tiny{$\pm 0.38$} } & { 10.09\tiny{$\pm 0.74$} } & { 13.59\tiny{$\pm 0.51$} }\\
		WPZ+GloVe & { 17.92\tiny{$\pm 0.05$} } & { 22.37\tiny{$\pm 0.11$} } & { 19.90\tiny{$\pm 0.08$} } & { 42.61\tiny{$\pm 0.08$} } & { 22.30\tiny{$\pm 0.03$} } & { 22.79\tiny{$\pm 0.05$} } & { 16.75\tiny{$\pm 0.08$} } & { 23.52\tiny{$\pm 0.07$} }\\
		WPZ+BERT & { 46.72\tiny{$\pm 1.03$} } & { 40.07\tiny{$\pm 0.48$} } & { 50.78\tiny{$\pm 2.09$} } & { 68.73\tiny{$\pm 1.87$} } & { 60.81\tiny{$\pm 1.70$} } & { 55.58\tiny{$\pm 3.56$} } & { 67.67\tiny{$\pm 1.16$} } & { 55.77\tiny{$\pm 1.70$} }\\
		\midrule
		TapNet & { 51.12\tiny{$\pm 5.36$} } & { 40.65\tiny{$\pm 2.83$} } & { 48.41\tiny{$\pm 2.27$} } & { 77.50\tiny{$\pm 1.09$} } & { 49.77\tiny{$\pm 1.36$} } & { 54.79\tiny{$\pm 2.32$} } & { 61.39\tiny{$\pm 2.41$} } & { 54.80\tiny{$\pm 2.52$} }\\
		TapNet+CDT & { 66.30\tiny{$\pm 3.81$} } & { 55.93\tiny{$\pm 1.78$} } & { 57.55\tiny{$\pm 6.57$} } & { 83.32\tiny{$\pm 0.96$} } & { 64.45\tiny{$\pm 4.07$} } & { 65.65\tiny{$\pm 1.74$} } & { 67.91\tiny{$\pm 3.32$} } & { 65.87\tiny{$\pm 3.18$} }\\
		L-WPZ+CDT & { 71.23\tiny{$\pm 6.00$} } & { 47.38\tiny{$\pm 4.18$} } & { 59.57\tiny{$\pm 5.55$} } & { 81.98\tiny{$\pm 2.08$} } & { 69.83\tiny{$\pm 1.94$} } & { 66.52\tiny{$\pm 2.72$} } & { 62.84\tiny{$\pm 0.58$} } & { 65.62\tiny{$\pm 3.29$} }\\
		L-TapNet+CDT & { \textbf{71.53}\tiny{$\pm 4.04$} } & { \textbf{60.56}\tiny{$\pm 0.77$} } & { \textbf{66.27}\tiny{$\pm 2.71$} } & { \textbf{84.54}\tiny{$\pm 1.08$} } & { \textbf{76.27}\tiny{$\pm 1.72$} } & { \textbf{70.79}\tiny{$\pm 1.60$} } & { 62.89\tiny{$\pm 1.88$} } & { \textbf{70.41}\tiny{$\pm 1.97$} }\\
		
		\bottomrule
	\end{tabular}
	}
	\caption{ \footnotesize  
		 1-shot slot tagging results with standard deviations.
	}\label{tbl:1shot_std}
\end{table*}

\begin{table*}[t]
	\centering
	\scriptsize
	\resizebox{\linewidth}{!}{
	\begin{tabular}{lrrrrrrrrr}\toprule
		\multirow{2}{*}{\textbf{Model}} &
		\multicolumn{7}{c}{\textbf{5-shots Slot Tagging}} & \\
		\cmidrule(lr){2-8}
		& \multicolumn{1}{c}{\textbf{We}} & \multicolumn{1}{c}{\textbf{Mu}} & \multicolumn{1}{c}{\textbf{Pl}} & \multicolumn{1}{c}{\textbf{Bo}} & \multicolumn{1}{c}{\textbf{Se}} & \multicolumn{1}{c}{\textbf{Re}} & \multicolumn{1}{c}{\textbf{Cr}} & \multicolumn{1}{c}{\textbf{Ave.}} \\
		\midrule
		
		Bi-LSMT & { 25.17\tiny{$\pm 0.42$} } & { 39.80\tiny{$\pm 0.52$} } & { 46.13\tiny{$\pm 0.42$} } & { 74.60\tiny{$\pm 0.21$} } & { 53.47\tiny{$\pm 0.45$} } & { 40.35\tiny{$\pm 0.52$} } & { 25.10\tiny{$\pm 0.94$} } & { 43.52\tiny{$\pm 0.50$} }\\
		SimBERT & { 53.46\tiny{$\pm 0.00$} } & { 54.13\tiny{$\pm 0.00$} } & { 42.81\tiny{$\pm 0.00$} } & { 75.54\tiny{$\pm 0.00$} } & { 57.10\tiny{$\pm 0.00$} } & { 55.30\tiny{$\pm 0.00$} } & { 32.38\tiny{$\pm 0.00$} } & { 52.96\tiny{$\pm 0.00$} }\\
		TransferBERT & { 59.41\tiny{$\pm 0.30$} } & { 42.00\tiny{$\pm 2.83$} } & { 46.07\tiny{$\pm 4.32$} } & { 20.74\tiny{$\pm 3.36$} } & { 28.20\tiny{$\pm 0.29$} } & { 67.75\tiny{$\pm 1.28$} } & { 58.61\tiny{$\pm 3.67$} } & { 46.11\tiny{$\pm 2.29$} }\\
		MN & { 36.67\tiny{$\pm 3.64$} } & { 33.67\tiny{$\pm 6.12$} } & { 52.60\tiny{$\pm 2.84$} } & { 69.09\tiny{$\pm 2.36$} } & { 38.42\tiny{$\pm 4.06$} } & { 33.28\tiny{$\pm 2.99$} } & { 72.10\tiny{$\pm 1.48$} } & { 47.98\tiny{$\pm 3.36$} }\\
		WPZ & { 9.54\tiny{$\pm 0.19$} } & { 14.23\tiny{$\pm 0.19$} } & { 18.12\tiny{$\pm 1.41$} } & { 44.65\tiny{$\pm 2.58$} } & { 18.98\tiny{$\pm 0.58$} } & { 12.03\tiny{$\pm 0.58$} } & { 14.05\tiny{$\pm 0.63$} } & { 18.80\tiny{$\pm 0.88$} }\\
		WPZ+GloVe & { 26.61\tiny{$\pm 0.54$} } & { 34.25\tiny{$\pm 0.16$} } & { 22.11\tiny{$\pm 0.04$} } & { 50.55\tiny{$\pm 0.15$} } & { 28.53\tiny{$\pm 0.05$} } & { 34.16\tiny{$\pm 0.43$} } & { 23.69\tiny{$\pm 0.07$} } & { 31.41\tiny{$\pm 0.21$} }\\
		WPZ+BERT & { 67.82\tiny{$\pm 4.11$} } & { 55.99\tiny{$\pm 2.24$} } & { 46.02\tiny{$\pm 3.19$} } & { 72.17\tiny{$\pm 1.75$} } & { 73.59\tiny{$\pm 1.60$} } & { 60.18\tiny{$\pm 6.96$} } & { 66.89\tiny{$\pm 2.88$} } & { 63.24\tiny{$\pm 3.25$} }\\
		\midrule
		TapNet & { 53.03\tiny{$\pm 7.20$} } & { 49.80\tiny{$\pm 3.02$} } & { 54.90\tiny{$\pm 2.72$} } & { 83.36\tiny{$\pm 1.03$} } & { 63.07\tiny{$\pm 1.96$} } & { 59.84\tiny{$\pm 1.57$} } & { 67.02\tiny{$\pm 2.51$} } & { 61.57\tiny{$\pm 2.86$} }\\
		TapNet+CDT & { 66.48\tiny{$\pm 4.09$} } & { 66.36\tiny{$\pm 1.77$} } & { 68.23\tiny{$\pm 3.99$} } & { \textbf{85.76}\tiny{$\pm 1.65$} } & { 73.60\tiny{$\pm 1.09$} } & { 64.20\tiny{$\pm 4.99$} } & { 68.47\tiny{$\pm 1.93$} } & { 70.44\tiny{$\pm 2.79$} }\\
		L-WPZ+CDT & { \textbf{74.68}\tiny{$\pm 2.43$} } & { 56.73\tiny{$\pm 3.23$} } & { 52.20\tiny{$\pm 3.22$} } & { 78.79\tiny{$\pm 2.11$} } & { 80.61\tiny{$\pm 2.27$} } & { 69.59\tiny{$\pm 2.78$} } & { 67.46\tiny{$\pm 1.91$} } & { 68.58\tiny{$\pm 2.56$} }\\
		L-TapNet+CDT & { 71.64\tiny{$\pm 3.62$} } & { \textbf{67.16}\tiny{$\pm 2.97$} } & { \textbf{75.88}\tiny{$\pm 1.51$} } & { 84.38\tiny{$\pm 2.81$} } & { \textbf{82.58}\tiny{$\pm 2.12$} } & { \textbf{70.05}\tiny{$\pm 1.61$} } & { \textbf{73.41}\tiny{$\pm 2.61$} } & { \textbf{75.01}\tiny{$\pm 2.46$} }\\
		\bottomrule
	\end{tabular}
	}
	\caption{ \footnotesize 
		5-shot slot tagging results with standard deviations.
	}\label{tbl:5shot_std}
\end{table*}

\end{document}